\documentclass[article]{acmart}
\usepackage{color}
\AtBeginDocument{%
  }

\renewcommand\footnotetextcopyrightpermission[1]{}

\begin{document}

\title{Video Emotion Open-vocabulary Recognition Based on Multimodal Large Language Model}

\author{Mengying Ge}
\email{gemengying@kanzhun.com}
\affiliation{%
  \institution{BOSS ZhiPin}
  \city{Beijing}
  \country{China}
}
\author{Dongkai Tang}
\email{tangdongkai@kanzhun.com}
\affiliation{%
  \institution{BOSS ZhiPin}
  \city{Beijing}
  \country{China}
}
\author{Mingyang Li}
\email{limingyang01@kanzhun.com}
\affiliation{%
  \institution{BOSS ZhiPin}
  \city{Beijing}
  \country{China}
}

\renewcommand{\shortauthors}{Mengying Ge et al.}
\begin{abstract}
Multimodal emotion recognition is a task of great concern. However, traditional data sets are based on fixed labels, resulting in models that often focus on main emotions and ignore detailed emotional changes in complex scenes. This report introduces the solution of using MLLMs technology to generate open-vocabulary emotion labels from a video. The solution includes the use of framework, data generation and processing, training methods, results generation and multi-model co-judgment. In the MER-OV (Open-Word Emotion Recognition) of the MER2024 challenge, our method achieved significant advantages, leading to its superior capabilities in complex emotion computation.
\end{abstract}



\maketitle

\section{Introduction}
Multimodal emotion recognition technology occupies a pivotal position in the field of human-computer interaction. This technology is committed to accurately capturing and identifying people's complex emotional states by integrating multiple modal information such as visual, auditory and textual semantics\cite{picard2000affective}.Currently, most research in this field focuses on model construction based on fixed label data sets. Although the objectivity of labels is strived for through a multi-person voting mechanism, the subjectivity and diversity of emotions themselves make it difficult for a single vocabulary to fully depict the full picture of individual emotions.

With the rapid advancement of large language models (LLMs) technology, many open problems have been solved unprecedentedly. However, in the field of multimodal emotion recognition, the application research of such technologies is still insufficient. Cutting-edge research such as MER2024\cite{lian2024mer} has evaluated the sentiment analysis capabilities of MLLMs such as Video-LLaMA\cite{zhang2023video}, SALMONN\cite{tang2023salmonn}, mPLUG-Owl\cite{ye2023mplug}, Qwen-Audio\cite{chu2023qwen}, and GPT-4V\cite{yang2023dawn}. These models have shown performance that exceeds heuristic baselines, demonstrating their potential in emotion understanding. In particular, the introduction of AffectGPT\cite{lian2024affectgpt} is not only committed to the accurate prediction of emotions, but also provides reasonable explanations behind the predictions, promoting the development of interpretable multimodal emotion reasoning.

This report explores in depth some research on multimodal emotion recognition using MLLMs technology, covering the optimization of infrastructure selection, refined data generation and processing processes, efficient training methods, and innovative solutions for result generation. The core research highlights are summarized as follows:
\begin{itemize}
\item \textbf{Emotion recognition training based on InternVL framework}. In view of the excellent performance of InternVL-Chat-V1.5 (hereinafter referred to as InternVL)\cite{chen2024far} in cross-domain tasks, we conducted in-depth fine-tuning on this basis. By using the generated character emotion description data and performing lora fine-tuning, InternVL's ability to parse character expressions was significantly enhanced. Experimental results show that this refined training significantly improves the performance of the model in emotion recognition tasks.
\item \textbf{Research on trimodal open vocabulary sentiment recognition}. The AffectGPT framework innovatively proposed a sentiment clue analysis framework that aligns the three modalities of image, speech, and text. We verified its superiority in open vocabulary sentiment recognition through SFT in downstream tasks, demonstrating the unique advantages of trimodal fusion in complex sentiment analysis.
\item \textbf{Synergy between MLLMs and traditional discriminative models}. Although traditional discriminative multimodal emotion recognition models can accurately capture the main emotion labels, they are slightly insufficient in capturing subtle emotions. MLLMs are good at capturing these subtle changes. Therefore, we explored a synergistic strategy that combines the advantages of both, aiming to achieve a more comprehensive and accurate judgment of character emotions through complementary effects.
\end{itemize}

\section{Proposed Method}
\subsection{InternVL Finetuning}

InternVL, an open-source multimodal large language model released by Shanghai Artificial Intelligence Laboratory, has shown excellent performance (SOTA) on the public test set, so we chose it as a benchmark and deeply explored its capabilities in both zero-shot and fine-tune modes.

In the zero-shot scenario, we directly use the open-source InternVL model without any task-specific training. Given that InternVL supports multi-image input, we frame the video and divide it into six parts, randomly selecting one frame from each part, and finally using these six frames as input. During zero-shot reasoning, we designed a specific prompt format: "These pictures are different frames of the same video. The words spoken by the characters in the picture are \{text\}. Assuming that you are an expert in the field of emotion, please describe the expression of the character in the picture in detail, and based on the above description, use a few words to summarize his expression in the format of [,,**]". \{text\} is the text content obtained by speech recognition .

In the fine-tune stage, we fine-tuned InternVL for downstream tasks based on the Swift framework\cite{reis2005swift}. Given that the verification results on the MER2024-SEMI track show that a large amount of human-centric data has a significant improvement on the emotion recognition task, we also adopted a similar strategy on the MER2024-OV track, that is, fine-tuning on a human-centric dataset. However, due to the scarcity of training data, we cleverly used the generation capabilities of Qwen-VL\cite{bai2023qwen} and CogVLM\cite{wang2023cogvlm} to create more captions data, and then used InternVL itself to screen these generated results to ensure the quality and relevance of the training data. This innovative data augmentation method effectively alleviates the problem of insufficient training data and improves the generalization ability of the model.

\subsection{Experiments Based on AffectGPT}
In the field of multimodal emotion recognition, the accuracy of emotion prediction can be significantly improved by integrating information from multiple modalities such as video, audio, and text for comprehensive judgment. The study of MERBench\cite{lian2024merbench} profoundly reveals the key role of the audio branch in discriminative emotion recognition methods. Given that most MLLMs mainly focus on understanding images or videos and text, AffectGPT innovatively introduces an audio branch based on the VideoLLava framework, aiming to comprehensively summarize the emotional state of the characters in the video by deeply analyzing the clues of these three modalities.

We designed the experiment based on the AffectGPT framework, using 332 description data provided by the official MER2024 competition, and scientifically divided them into a training set (accounting for 3/4, i.e., 266 samples) and a test set (accounting for 1/4, i.e., 66 samples). In the training stage, we used these training data to perform sft (soft fine-tuning) optimization training on the model, and guided the model through the following carefully designed prompts:
\textcolor{red}{"\#\#\#Human: Close your eyes, open your ears and you imagine only based on the sound that <Audio><AudioHere></Audio>. "Close your ears, open your eyes and you see that <Video><ImageHere></Video>. The subtitle content of this video is <Subtitle>{subtitle}</Subtitle>. Now as an expert in the field of emotions, please focus on the facial expressions, body movements, environment, acoustic information, subtitle content, etc., in the video to discern clues related to the emotions of the individual. Please provide a detailed description and ultimately predict the emotional state of the individual in the video. \#\#\#Assistant:"}

During the testing phase, we applied our trained model to perform predictive inference on 66 test set samples. The model analyzed the emotional states of individuals in videos from three independent perspectives: video, audio, and text. Finally, to obtain an open vocabulary list for emotion descriptions, we utilized the InternVL large language model to refine and summarize the analysis results at the lexical level, ultimately generating model outputs with rich emotional vocabularies that provide comprehensive and precise analyses for emotion recognition.

\subsection{Multi-Model Co-judgment}
When the accuracy of a single model reaches a bottleneck and is difficult to further improve, a multi-model integration strategy becomes a natural choice. In the context of the MER2024-OV competition, we also followed this idea. As representatives of large visual multi-modal models, AffectGPT and InternVL have the advantage of overall understanding and reasoning capabilities of video content, and can generally capture and summarize the expression information appearing in the video. However, in comparison, traditional multi-modal algorithms show higher accuracy on specific prediction tasks.
In view of this, we cleverly combined the strengths of the two and spliced and fused the prediction results of the multi-modal model in the MER2024-SEMI track, the output of AffectGPT, and the predictions of InternVL, aiming to improve the performance through the complementarity between them. Overall precision and recall. Experimental results show that this strategy significantly improves the overall evaluation index and verifies the effectiveness of multi-model integration in improving model performance.

\section{Experiments and Analysis}
\subsection{Dataset}
\subsubsection{Dataset Partitioning}

\begin{table}[H]
\caption{\textbf{Dataset Distribution}}
\centering
\resizebox{0.4\linewidth}{!}{
\begin{tabular}{ccc}%
\toprule
type & dataset & count \\
\midrule
train &MiniGPT-4 & 3439 \\
\midrule
train & MER2024-OV & 266 \\
\midrule%
test & MER2024-OV & 66 \\
\bottomrule
\end{tabular}}
\end{table}
Given the high cost of producing video, audio, and text emotion data, the MER2024 competition only provided 322 emotion description data to the contestants. In order to enhance the training effect and stability of the model, during the training process, we not only made full use of the 322 officially provided emotion description data, but also introduced the general text and image dataset MiniGPT-4\cite{zhu2023minigpt} as supplementary training materials. In addition, in order to verify the generalization ability of the model, we randomly selected 66 of the 322 data as a test set to evaluate the performance of the model in different scenarios. The specific distribution of the data is shown in Table 1.
\subsubsection{Dataset Generating}
Since generative large models usually require paired Image-Text data formats during fine-tuning, and our fine-tuning goal focuses on face-centered sentiment analysis, there is a lack of a large number of high-quality datasets on facial expressions suitable for large model training in existing open source datasets. Therefore, we adopted the method of generating data to make up for this deficiency. The main process is to use the capabilities of Qwen-VL and CogVLM to generate captions (descriptive text) that meet the needs. When generating captions, we carefully designed the prompt, "As an expert in the field of emotions, pay close attention to the facial expressions, body movements, environment, and subtitle content of the characters in the image to capture clues closely related to personal emotions, and provide detailed descriptions based on this, and finally predict the emotional state of the characters in the image." This design is intended to ensure that the generated caption can fully and accurately reflect the emotional information in the image. Subsequently, in order to improve the quality of training data, we used InternVL to compare text similarity. Specifically, for each pair of generated captions, we input them into InternVL and set the prompt to: "Please judge whether the emotions described in these two sentences are similar and give a score between 0 and 1." Through this step, we can effectively eliminate low-quality captions with a similarity lower than 0.9, and for caption pairs with a similarity higher than 0.9, we randomly select one of them to join the training set.

Finally, we successfully generated 26,000 pairs of high-quality captions based on the open source dataset CH-SIMS-v2\cite{yu2020ch}. These captions will be used as the final training set to improve the performance of our model in facial emotion analysis.

\subsection{Results and Analysis}
\subsubsection{Settings}
In the fine-tuning training of AffectGPT, we focused on optimizing its Audio Q-Former and Video Q-Former layers. The training process used 4 A800 GPUs with a batch size of 4 and lasted for 100 training cycles (epochs), which took a total of about 33 hours.
For the fine-tuning of InternVL, we relied on the Swift framework and used lora (Low-Rank Adaptation) technology to fine-tune parameters to reduce the amount of calculation while maintaining model performance. The training dataset consists of 26,000 carefully constructed image-text pairs. During the training process, 8 A800 GPUs were used for acceleration, and the batch size was 1 to stabilize the training process. In addition, we also set the maximum input length to 8192 to support the processing of more complex image and text information.

\subsubsection{Analysis}
In the experiment on the InternVL model, we conducted a comparative evaluation between the zero-shot test and lora fine-tuning based on the 322 officially provided data sets. The core measurement standard of the experiment is Avg score. The specific definition of this indicator can refer to the MER2024 specification. From the experimental results Table 1, we observed that after lora fine-tuning, the Avg score significantly increased by 3\%, which strongly supports our hypothesis: fine-tuning on a human-centered data set can effectively promote the model's performance in downstream tasks.

Furthermore, when delving deeper into the inference process, we noticed the impact of face region preprocessing on the final performance. Through comparative experiments, we found that the Avg score is 11\% higher than using the entire image as input for prediction compared to the strategy of first cutting out the face and performing alignment processing. This finding may be attributed to the fact that after the face is cut out, some detailed information is lost due to the reduction in resolution, thus affecting the prediction accuracy of the model.
\begin{table}[H]
\caption{\textbf{Experimental Results}}
\centering
\resizebox{0.8\linewidth}{!}{
\begin{tabular}{ccccc}%
\toprule
model & preprocess & avg &accuracy &recall \\
\midrule
InternVL-Chat-V1.5(zero-shot) &face alignment& 0.3292 & 0.2001 & 0.4582 \\
\midrule
InternVL-Chat-V1.5(zero-shot) &entire image& 0.4394 & 0.3170 & 0.5618 \\
\midrule%
InternVL-Chat-V1.5(fine-tune) &entire image& 0.4701 & 0.4222 & 0.5179 \\
\midrule
InternVL-Chat-V1.5(zero-shot+fine-tune) &entire image& 0.5258 & 0.3503 & 0.7013 \\
\midrule
InternVL-Chat-V1.5(zero-shot+fine-tune) + discriminative model &entire image& \textbf{0.6167} & 0.3895 & 0.8439 \\
\bottomrule
\end{tabular}}
\end{table}
In order to further improve model performance, we also tried a multi-model integration strategy. Specifically, we integrated the zero-shot results of InternVL, the results after lora fine-tuning, and the output of the discriminative model based on this. As can be seen from the data in Table 1, although the integrated model has a slight decrease in accuracy, it significantly improves the Recall value, which in turn promotes the increase in the overall Avg score. This shows that through a reasonable model integration strategy, we can effectively improve the recall rate while maintaining a certain accuracy, thereby optimizing the overall performance.

Table 3 shows the performance of AffectGPT on a 66-word open vocabulary test set after fine-tuning. The results prove that the framework can effectively acquire the ability to understand complex emotions with only a small amount of SFT training data, and shows excellent generalization performance on the test set.
\begin{table}[H]
\caption{\textbf{AffectGPT SFT Results}}
\centering
\resizebox{0.4\linewidth}{!}{
\begin{tabular}{cccc}%
\toprule
model & avg &accuracy &recall \\
\midrule
AffectGPT-sft & 0.7429 & 0.6775 & 0.8083 \\
\bottomrule
\end{tabular}}
\end{table}
\section{Conclusion and Limitations}
This report primarily describes the technical approach we used in the MER2024-OV track. We mainly fine-tuned InternVL on human-related data, which helps the model better understand and extract facial expressions from details such as facial expressions, body movements, and surrounding environments. Additionally, we leveraged the powerful multimodal capabilities of AffectGPT, which integrates speech, vision, and text, and performed fine-tuning on small batches of data, which also had a positive impact on the entire task. Finally, we integrated multiple models to improve recall, and the final results indicate that our methods provided a significant advantage.

However, this approach also has limitations. Model integration, while improving recall, reduced precision. The reason is that large models, while capable of understanding primary expressions, can also introduce many irrelevant or secondary expressions, leading to a decrease in accuracy. For future work, we can consider how to reduce the output of invalid expressions from large models, retaining only a few critical expressions.

\bibliographystyle{unsrt}
\bibliography{references.bib}

\begin{thebibliography}{10}

\bibitem{picard2000affective}
Rosalind~W Picard.
\newblock {\em Affective computing}.
\newblock MIT press, 2000.

\bibitem{lian2024mer}
Zheng Lian, Haiyang Sun, Licai Sun, Zhuofan Wen, Siyuan Zhang, Shun Chen, Hao Gu, Jinming Zhao, Ziyang Ma, Xie Chen, et~al.
\newblock Mer 2024: Semi-supervised learning, noise robustness, and open-vocabulary multimodal emotion recognition.
\newblock {\em arXiv preprint arXiv:2404.17113}, 2024.

\bibitem{zhang2023video}
Hang Zhang, Xin Li, and Lidong Bing.
\newblock Video-llama: An instruction-tuned audio-visual language model for video understanding.
\newblock {\em arXiv preprint arXiv:2306.02858}, 2023.

\bibitem{tang2023salmonn}
Changli Tang, Wenyi Yu, Guangzhi Sun, Xianzhao Chen, Tian Tan, Wei Li, Lu~Lu, Zejun Ma, and Chao Zhang.
\newblock Salmonn: Towards generic hearing abilities for large language models.
\newblock {\em arXiv preprint arXiv:2310.13289}, 2023.

\bibitem{ye2023mplug}
Qinghao Ye, Haiyang Xu, Guohai Xu, Jiabo Ye, Ming Yan, Yiyang Zhou, Junyang Wang, Anwen Hu, Pengcheng Shi, Yaya Shi, et~al.
\newblock mplug-owl: Modularization empowers large language models with multimodality.
\newblock {\em arXiv preprint arXiv:2304.14178}, 2023.

\bibitem{chu2023qwen}
Yunfei Chu, Jin Xu, Xiaohuan Zhou, Qian Yang, Shiliang Zhang, Zhijie Yan, Chang Zhou, and Jingren Zhou.
\newblock Qwen-audio: Advancing universal audio understanding via unified large-scale audio-language models.
\newblock {\em arXiv preprint arXiv:2311.07919}, 2023.

\bibitem{yang2023dawn}
Zhengyuan Yang, Linjie Li, Kevin Lin, Jianfeng Wang, Chung-Ching Lin, Zicheng Liu, and Lijuan Wang.
\newblock The dawn of lmms: Preliminary explorations with gpt-4v (ision).
\newblock {\em arXiv preprint arXiv:2309.17421}, 9(1):1, 2023.

\bibitem{lian2024affectgpt}
Zheng Lian, Haiyang Sun, Licai Sun, Jiangyan Yi, Bin Liu, and Jianhua Tao.
\newblock Affectgpt: Dataset and framework for explainable multimodal emotion recognition.
\newblock {\em arXiv preprint arXiv:2407.07653}, 2024.

\bibitem{chen2024far}
Zhe Chen, Weiyun Wang, Hao Tian, Shenglong Ye, Zhangwei Gao, Erfei Cui, Wenwen Tong, Kongzhi Hu, Jiapeng Luo, Zheng Ma, et~al.
\newblock How far are we to gpt-4v? closing the gap to commercial multimodal models with open-source suites.
\newblock {\em arXiv preprint arXiv:2404.16821}, 2024.

\bibitem{reis2005swift}
George~A Reis, Jonathan Chang, Neil Vachharajani, Ram Rangan, and David~I August.
\newblock Swift: Software implemented fault tolerance.
\newblock In {\em International symposium on Code generation and optimization}, pages 243--254. IEEE, 2005.

\bibitem{bai2023qwen}
Jinze Bai, Shuai Bai, Shusheng Yang, Shijie Wang, Sinan Tan, Peng Wang, Junyang Lin, Chang Zhou, and Jingren Zhou.
\newblock Qwen-vl: A frontier large vision-language model with versatile abilities.
\newblock {\em arXiv preprint arXiv:2308.12966}, 2023.

\bibitem{wang2023cogvlm}
Weihan Wang, Qingsong Lv, Wenmeng Yu, Wenyi Hong, Ji~Qi, Yan Wang, Junhui Ji, Zhuoyi Yang, Lei Zhao, Xixuan Song, et~al.
\newblock Cogvlm: Visual expert for pretrained language models.
\newblock {\em arXiv preprint arXiv:2311.03079}, 2023.

\bibitem{lian2024merbench}
Zheng Lian, Licai Sun, Yong Ren, Hao Gu, Haiyang Sun, Lan Chen, Bin Liu, and Jianhua Tao.
\newblock Merbench: A unified evaluation benchmark for multimodal emotion recognition.
\newblock {\em arXiv preprint arXiv:2401.03429}, 2024.

\bibitem{zhu2023minigpt}
Deyao Zhu, Jun Chen, Xiaoqian Shen, Xiang Li, and Mohamed Elhoseiny.
\newblock Minigpt-4: Enhancing vision-language understanding with advanced large language models.
\newblock {\em arXiv preprint arXiv:2304.10592}, 2023.

\bibitem{yu2020ch}
Wenmeng Yu, Hua Xu, Fanyang Meng, Yilin Zhu, Yixiao Ma, Jiele Wu, Jiyun Zou, and Kaicheng Yang.
\newblock Ch-sims: A chinese multimodal sentiment analysis dataset with fine-grained annotation of modality.
\newblock In {\em Proceedings of the 58th annual meeting of the association for computational linguistics}, pages 3718--3727, 2020.

\end{thebibliography}
\end{document}